\documentclass[11pt]{article}

\usepackage{acl}

\usepackage{amsmath,amsfonts,bm}

\def\eqref#1{equation~\ref{#1}}

\def\1{\bm{1}}

\def\vb{{\bm{b}}}

\def\vh{{\bm{h}}}

\def\vx{{\bm{x}}}

\def\mP{{\bm{P}}}

\def\mW{{\bm{W}}}

\DeclareMathAlphabet{\mathsfit}{\encodingdefault}{\sfdefault}{m}{sl}
\SetMathAlphabet{\mathsfit}{bold}{\encodingdefault}{\sfdefault}{bx}{n}

\newcommand{\R}{\mathbb{R}}

\DeclareMathOperator*{\argmax}{arg\,max}

\usepackage{times}
\usepackage{latexsym}

\usepackage{CJKutf8}
\usepackage{booktabs}
\usepackage{graphicx}
\usepackage{subfigure}
\usepackage{comment}
\usepackage{xspace}
\usepackage{adjustbox}
\usepackage{mathtools}
\usepackage{bbm}
\usepackage{multirow}

\usepackage[T1]{fontenc}

\usepackage[utf8]{inputenc}

\usepackage{microtype}
\usepackage{xcolor}
\usepackage{fancyhdr}
\title{MoE\emph{fication}: Transformer Feed-forward Layers are Mixtures of Experts}

\author{Zhengyan Zhang$^{1,2}$, Yankai Lin$^{3}$, Zhiyuan Liu$^{1,2,4,5\dagger}$, \\
\textbf{Peng Li$^{3,6}$, Maosong Sun$^{1,2,4,5,7\dagger}$, Jie Zhou$^{3}$}\\
        $^1$Dept. of Comp. Sci. \& Tech., Institute for AI, Tsinghua University, Beijing, China
        \\
         $^2$Beijing National Research Center for Information Science and Technology\\
         $^3$Pattern Recognition Center, WeChat AI, Tencent Inc \\ 
         $^4$International Innovation Center of Tsinghua University, Shanghai, China \\
         $^5$Beijing Academy of Artificial Intelligence \\
         $^6$Institute for AI Industry Research (AIR), Tsinghua University, China\\
         $^7$Jiangsu Collaborative Innovation Center for Language Ability, Xuzhou, China \\
         \texttt{zy-z19@mails.tsinghua.edu.cn} \texttt{\{liuzy,sms\}@tsinghua.edu.cn}\\
}

\begin{document}
\maketitle
\begin{abstract}


Recent work has shown that feed-forward networks (FFNs) in pre-trained Transformers are a key component, storing various linguistic and factual knowledge. However, the computational patterns of FFNs are still unclear. In this work, we study the computational patterns of FFNs and observe that most inputs only activate a tiny ratio of neurons of FFNs. This phenomenon is similar to the sparsity of the human brain, which drives research on functional partitions of the human brain. To verify whether functional partitions also emerge in FFNs, we propose to convert a model into its MoE version with the same parameters, namely MoE\emph{fication}. Specifically, MoEfication consists of two phases: (1) splitting the parameters of FFNs into multiple functional partitions as experts, and (2) building expert routers to decide which experts will be used for each input. Experimental results show that MoEfication can conditionally use $10\%$ to $30\%$ of FFN parameters while maintaining over $95\%$ original performance for different models on various downstream tasks. Besides, MoEfication brings two advantages: (1) it significantly reduces the FLOPS of inference, i.e., 2x speedup with $25\%$ of FFN parameters, and (2) it provides a fine-grained perspective to study the inner mechanism of FFNs. The source code of this paper can be obtained from \url{https://github.com/thunlp/MoEfication}. 
\end{abstract}

\section{Introduction}

{\let\thefootnote\relax\footnotetext{$^\dagger$ Corresponding authors}}
{\let\thefootnote\relax\footnotetext{Part of the work was done while Peng Li was working at Tencent.}}


Recent years have witnessed great success of Transformer-based pre-trained language models (PLMs)~\cite{BERT,GPT-3,Han2021-yr}, attracting many efforts to interpret the inner mechanism of Transformer~\cite{DBLP:journals/pnas/ManningCHKL20,kovaleva-etal-2019-revealing}. However, most of these works focus on the attention mechanism but ignore the feed-forward networks (FFNs), which constitute nearly two-thirds of model parameters. Although recent work has shown that FFNs can be viewed as memory networks storing amounts of knowledge~\cite{geva-etal-2021-transformer,dai2021knowledge}, the computational patterns of FFNs are still unclear.

In this work, we study the activation patterns of FFNs in Transformer models and find a phenomenon of \textbf{sparse activation}, i.e., only a tiny fraction of neurons are activated for a single input. For example, when we perform inference on a fine-tuned T5-Large model~\cite{T5} with $700$-million parameters, $90\%$ inputs only activate less than $5\%$ neurons\footnote{T5 uses ReLU as the activation function. We treat the neurons having positive outputs as activated neurons.}. This phenomenon is similar to the sparsity in the human brain~\cite{olshausen1996emergence,gross2002genealogy}, which drives research on functional partitions of the human brain~\cite{garey1999brodmann}. 
Inspired by such observation, we further raise up a question: do the functional partitions also emerge in artificial neural models, i.e., FFNs in pre-trained Transformer?

To investigate this problem, we explore whether a Transformer can be converted into an equivalent Mixture-of-Experts (MoE) model~\cite{ConditionalComputation}, which regards different functional partitions in FFNs as different experts conditionally activated. Specially, we propose MoEfication to discover the functional partitions (experts) in FFNs and build routers for selecting experts. It consists of two phases. (1) \textbf{Expert Construction}: Split a whole feed-forward layer into multiple experts. The goal is to group those neurons that are often activated simultaneously into the same expert network. (2) \textbf{Expert Selection}: Select those experts that contain as many activated neurons as possible for each input to approximate to the original results.

In the experiments, we evaluate MoEfication on two typical kinds of downstream tasks, including GLUE and QA benchmarks~\cite{GLUE,SQuAD,RACE}, using T5 and BERT~\cite{T5,BERT}.
Experimental results verify that FFNs in Transformers can be converted to mixtures of experts, and thus we can use only $10\%$ to $30\%$ of FFN parameters to maintain over $95\%$ original performance, which verifies that the pre-trained Transformers also learn the functional partitions in FFNs. 
Besides, MoEfication brings two advantages: (1) It can significantly speed up the inference of Transformers. Using $25\%$ of FFN parameters brings 2x speedup on CPU and 1.2x speedup on GPU. (2) We can study MoEfied models to interpret the inner mechanism of FFNs at a fine-grained level. In this work, we study their routing patterns and hope these findings can help future work on the design and training of MoE models.

\section{Related Work}



\textbf{Interpretation of Large-scale Transformers.} Due to the success of Transformer-based PLMs, there are many studies on the interpretation of Transformer, including the functionality of different layers~\cite{BERT-Pipeline,jawahar-etal-2019-bert,wang-tu-2020-rethinking,ramnath-etal-2020-towards}, and the mechanisms of both attention networks and FFNs~\cite{DBLP:journals/pnas/ManningCHKL20,kovaleva-etal-2019-revealing,wallace-etal-2019-allennlp}. Recent work find that the FFNs of Transformers can be viewed as memory networks storing lots of knowledge learned from language modeling~\cite{geva-etal-2021-transformer,dai2021knowledge,DBLP:journals/corr/abs-2005-07647}. Meanwhile, researchers explore to modify the knowledge stored in FFNs and achieve promising results~\cite{de-cao-etal-2021-editing,meng2022locating}. In this work, we show that how the knowledge stored in FFNs is used, that is, most FFNs can be viewed as a MoE network where the knowledge is conditionally activated.

\begin{figure*}
  \centering
  \includegraphics[width=\linewidth]{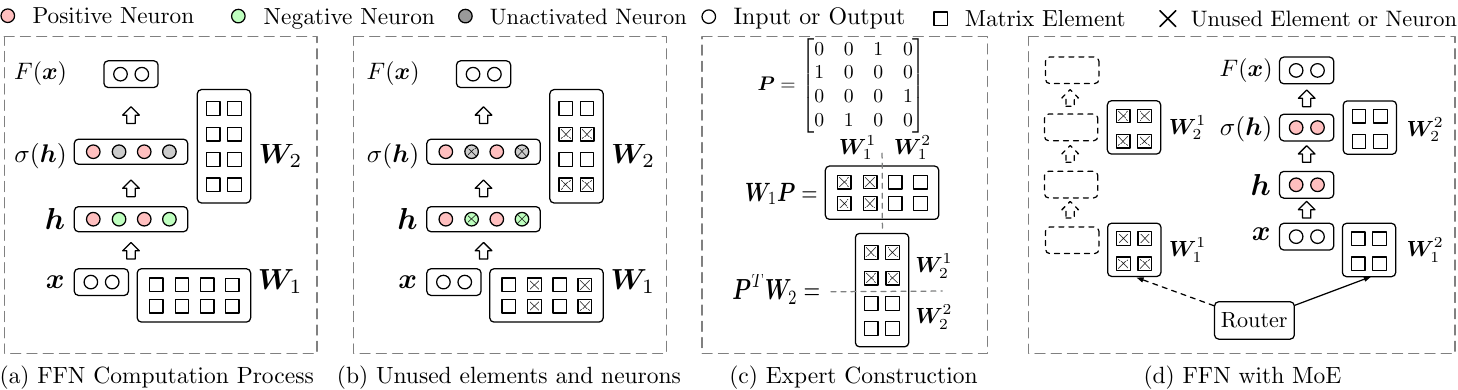}
  \caption{An example of the sparse activation phenomenon and MoEfication. (a) shows the computation process of an FFN for a given input. (b) shows the unused elements and neurons for this input. (c) shows how to construct experts. (d) shows how the MoEfied model handles this input efficiently.}
  \label{fig:model}
\end{figure*}

\textbf{Large-scale PLMs with MoE.} \citet{Jacobs1991-zn} propose mixture-of-experts to build a system composed of many separate networks, which learn to handle a subset of the training examples independently. When deep neural networks achieve great success~\cite{Hinton2012-lt,Krizhevsky2012-pw,goodfellow2013maxout}, \citet{ConditionalComputation} thinks the model size is a key factor and MoE is an important technique to scaling model computation and proposes the idea of ``conditional computation''. The first large-scale MoE language model is proposed by~\citet{moe-lstm}, which adds an MoE layer between two LSTM layers and independently assigns tokens to combinations of experts. Recently, GShard~\cite{GShard}, Switch-Transformer~\cite{Switch-Transformer}, BASELayer~\cite{Lewis2021-nh}, and HashLayer~\cite{hashlayer} study how to build large-scale Transformer-based models with MoE and optimal training strategies, which can fully utilize the model capacity. Different from them, we utilize the naturally-existing sparse activation phenomenon to convert a model into its MoE version for better efficiency during inference.

\textbf{Model Acceleration for PLMs.} Model acceleration aims to reduce the time and space complexity of PLMs. 
There are several techniques including knowledge distillation~\cite{DistillBERT,PKD,TinyBERT}, model pruning~\cite{Voita2019-sl,michel2019are,Zhang2021-wn}, attention approximation~\cite{wang2020linformer,DBLP:conf/iclr/KitaevKL20,DBLP:conf/nips/ZaheerGDAAOPRWY20},and model quantization~\cite{Zafrir2019-je,TernaryBERT,DBLP:conf/acl/BaiZHSJJLLK20}, and dynamic inference~\cite{DeeBERT,CascadeBERT,DBLP:conf/naacl/YeLHS21,DBLP:conf/nips/HouHSJCL20}. 
Among these techniques, dynamic inference explore to selectively omit unnecessary computation for acceleration, which is similar to the target of MoEfication. 
Previous work usually focuses on how to dynamically drop layers to accelerate inference~\cite{MSDNet,DynamicInference,CascadeBERT}, which introduces additional training objectives and prediction strategies. In contrast, MoEfication simplifies models in a finer granularity, and does not change the process of training and inference. In summary, MoEfication can be regarded as a novel direction diagonal with the above-mentioned approaches.



\section{MoEfication}

In this section, we will introduce the general idea of MoEfication and divide it into two phases: expert construction and expert selection.

\subsection{Overall Framework}

MoEfication aims to utilize the sparse activation phenomenon in the FFNs of Transformers to reduce the computation cost. 

We first formally describe the sparse activation phenomenon. The FFNs of Transformers are two-layer fully connected networks, which process an input representation $\vx \in \R^{d_{model}}$ by
\begin{equation}
  \small
  \begin{aligned}
    \vh &= \vx\mW_1+\vb_1, \\
    F(\vx) &= \sigma(\vh)\mW_2+\vb_2,
  \end{aligned}
\end{equation}
where $\mW_1 \in \R^{d_{model} \times d_{ff}}$ and $\mW_2 \in \R^{d_{ff} \times d_{model}}$ are the weight matrices, $\vb_1 \in \R^{d_{ff}}$ and $\vb_2 \in \R^{d_{model}}$ are the bias vectors, and $\sigma(\cdot)$ is a non-linear activation function, which prefers to retain positive values and discard negative ones. In this work, we study the activation function ReLU~\cite{ReLU}, which is used by the original Transformer~\cite{Transformer} and some widely-used Transformer-based PLMs~\cite{Sun2020-dr,T5}.


Since there are many inactive (zero) values in the intermediate output $\sigma(\vh)$, the computation of these values can be omitted for acceleration. Meanwhile, different inputs will activate different neurons. Hence, we explore to select the possiblely-activated neurons of $\vh$ before the FFN computation instead of model pruning.

We show an example in Figure~\ref{fig:model}. In this FFN, $d_{model}$ is $2$, $d_{ff}$ is $4$, and the bias vectors are omitted for simplification. For a given input representation $\vx$, there are two positive values in $\vh$. Hence, we only need to compute part of the FFN, i.e., a $2\times2$ submatrix of $\mW_1$ and a $2\times2$ submatrix of $\mW_2$, to obtain the same output $F(\vx)$. Correspondingly, we can MoEfy the original FFN to have an MoE layer with two experts and select the one on the right-hand side for this input $\vx$.

For MoEfication, we first split the FFN into several independent parts, namely expert construction, and then design a router to select suitable experts for each input, namely expert selection.

\subsection{Expert Construction}

In this subsection, we introduce how to split an FFN into several parts. The core idea is to group together the neurons that are often activated simultaneously. In this way, for each input, we can select a small number of experts to cover all its activated neurons. To achieve better parallel computation performance, we set the size of each expert to be the same. If the number of experts is $k$, the input and output dimension of experts is still $d_{model}$ and their intermediate dimension is $d_e = \frac{d_{ff}}{k}$. Then, the parameters of $i$-th expert are denoted by
\begin{equation}
  \small
  \mW_1^i \in \R^{d_{model} \times d_e}, \vb_1^i \in \R^{d_e}, \mW_2^i \in \R^{d_e\times d_{model}}.
\end{equation}

Given the result of splitting, we construct the corresponding permutation of intermediate neurons by $\bigl(\begin{smallmatrix}
    1 & 2 & \ldots & d_{ff} \\
    f(1) & f(2) & \ldots & f(d_{ff})
  \end{smallmatrix}\bigr)$, where $f(n)$ is the mapping function from the original neuron index to the permuted neuron index. We compute $f(n)$ by
\begin{equation}
    \small
    f(n) = (e(n)-1)d_e + |\{m|m\le n,e(m)=e(n)\}|,
\end{equation}
where $e(n)$ is the expert index of the $n$-th neuron, which varies from $1$ to $k$, and $|\{m|m\le n,e(m)=e(n)\}|$ is the index of the $n$-th neuron in the expert.
Then, we use its permutation matrix $\mP \in \R^{d_{ff}\times d_{ff}}$ to permute the rows or columns of parameters and have the following split:
\begin{equation}
\small
\begin{aligned}
  [\mW_1^1, \mW_1^2, \ldots, \mW_1^k] &= \mW_1\mP, \\
  \vb_1^1 \oplus \vb_1^2 \oplus \ldots \oplus \vb_1^k &= \vb_1\mP, \\
  [(\mW_2^1)^T, (\mW_2^2)^T, \ldots, (\mW_2^k)^T] &= (\mP^T\mW_2)^T, \\
\end{aligned}
\end{equation}
where $\oplus$ represents the vertical concatenation. Note that the permutation will not influence the output representation:
\begin{equation}
\small
\begin{aligned}
    \sigma(\vh)\mW_2+\vb_2 &= \sigma(\vh)\mP\mP^T\mW_2+\vb_2, \\
    &= \sigma(\vh\mP)\mP^T\mW_2+\vb_2, \\
    &= \sigma(\vx \underline{\mW_1\mP}+\underline{\vb_1\mP})\underline{\mP^T\mW_2}+\vb_2. \\
\end{aligned}
\end{equation}

In this work, we propose two methods to split an FFN into $k$ parts.



\textbf{Parameter Clustering Split}. To take the parameter information into consideration, we treat the columns of $\mW_1$ as a collection of vectors with $d_{model}$ dimension. Based on the intuition that the neurons with similar vectors will be activated simultaneously, we apply balanced K-Means~\cite{bkmeans} to the vector collection to obtain $k$ clusters to construct the mapping function. 

\textbf{Co-Activation Graph Split}. To directly use the information of co-activation, we construct a co-activation graph by counting co-activations of PLMs for the samples of the training set. Each neuron will be represented by a node in the graph, and the edge weight between two nodes are their co-activation values. The co-activation value is computed by
\begin{equation}
    \small
    \text{co-activation}(n,m) = \sum_\vx \vh_n^{(\vx)}\vh_m^{(\vx)}\mathbbm{1}_{\vh_n^{(\vx)}>0, \vh_m^{(\vx)} > 0},
\end{equation}
where $\vh_n^{(\vx)}$, $\vh_m^{(\vx)}$ are the $n$-th and the $m$-th neurons of $\vh$ for the input $\vx$ and $\mathbbm{1}_{\vh_n^{(\vx)}>0, \vh_m^{(\vx)}>0}$ indicates $\vh_n^{(\vx)}$ and $\vh_m^{(\vx)}$ are activated simultaneously. Then, we apply graph partitioning algorithms~\cite{metis} to the co-activation graph to obtain the split, where the internal connections for each group will be strong. Please refer to Appendix~\ref{sec:partition} for the details of the partitioning algorithm. It means that the neurons splitted into the same group are often activated simultaneously for the training samples.

\subsection{Expert Selection}

In this subsection, we introduce how to create a router for expert selection. 
An MoEfied FFN processed an input $\vx$ by
\begin{equation}
    \small
    F_m(\vx) = \sum_{i\in S} \sigma(\vx\mW_1^i+\vb_1^i)\mW_2^i+\vb_2,
\end{equation}
where $S$ is the set of the selected experts. If all experts are selected, we have $F_m(\vx)=F(\vx)$.
Considering that $\sigma(\vx\mW_1^i+\vb_1^i)\mW_2^i$ equals to $\mathbf{0}$ for most experts, we try to select $n$ experts, where $n < k$, minimize $||F_m(\vx) - F(\vx)||_2$. 
The selection methods will assign a score $s_i$ to each expert for the given input $\vx$ and select the experts with the $n$ highest scores by
\begin{equation}
    \small
    \label{eq:greedy}
    S = \argmax_{A\subset \{1, 2, \ldots, k\}, |A|=n} \sum_{i \in A} s_i.
\end{equation}

\textbf{Groundtruth Selection} for the intermediate output $\sigma(\vh)$. We can obtain the groundtruth selection, which minimizes $||\textrm{concat}(\{ f(\sigma(\vx\mW_1^i+\vb_1^i))\mathbbm{1}(i\in S)\})-\sigma(\vh)||_2$, by a greedy algorithm. $f$ is a padding function with zeros to match the dimension between $\sigma(\vx\mW_1^i+\vb_1^i)$ and $\sigma(\vh)$. We calculate the sum of positive values in each expert as $s_i$ and select experts using Equation~\ref{eq:greedy}. This selection should approximate to the lower bound of $||F_m(\vx) - F(\vx)||_2$. Correspondingly, its performance will approximate to the ideal performance of an MoEfied model. Meanwhile, it is intractable to directly optimize $||F_m(\vx) - F(\vx)||_2$ because there are too many possible combinations of experts.


\textbf{Similarity Selection.} To utilize the parameter information, we average all columns of $\mW_1^i$ and use it as the expert representation. Given an input $\vx$, we calculate the cosine similarity between the expert representation and $\vx$ as $s_i$.

\textbf{MLP Selection.} We train a multi-layer perceptron (MLP), which takes the $\vx$ as input and predicts the sum of positive values in each expert. Then, we use the prediction as $s_i$. This method tries to approximate to the performance of groundtruth selection.


\section{Experiment}

\subsection{Experimental Setups}

\textbf{Models and Hyperparameter.} We use four variants of T5~\cite{T5}, which are the 60-million-parameter T5-Small, the 200-million-parameter T5-Base, the 700-million-parameter T5-Large, and the 3-billion-parameter T5-XLarge. The non-linear activation function is ReLU~\cite{ReLU}. We use Adam as the optimizer and a learning rate of $10^{-6}$ for fine-tuning T5 models on downstream tasks. The batch size is set to $64$ and the number of epochs is set to $3$.

\textbf{Datasets.} We use several natural language understanding datasets to evaluate our models. We use SST-2~\cite{socher2013recursive}, MNLI-matched~\cite{williams2018broad}, and RACE~\cite{RACE} as the main evaluation datasets, which cover single-sentence classification, sentence-pair classification, and reading comprehension. We report the results on their development sets. We also report the results of MoEfication in other datasets in Appendix~\ref{sec:other} including the tasks in GLUE benchmark~\cite{GLUE} and SQuAD~\cite{SQuAD}.

\textbf{Expert Construction.} 
For balanced K-Means, we use an open-source implementation\footnote{\url{https://github.com/ndanielsen/Same-Size-K-Means}}. 
Besides Parameter Clustering Split and Co-activation Graph Split, we also implement Random Split as a naive baseline, which uses an identity matrix as $\mP$. 
For the number of neurons in each expert, if the number is small, there will be a lot of experts, making the routing computation cost high. Meanwhile, if the number is large, there will be more inactive neurons in each expert for a given input, which is harmful to the performance with the same amount of selected neurons. Hence, selecting the number of neurons in each expert is a trade-off between computation cost and accuracy. 
According to our pilot experiments, we set the number of neurons in each expert $d_e$ to $32$.
Correspondingly, the number of experts varies from $64$ to $512$ ($k=\frac{d_{ff}}{d_e}$) for different T5 variants. 
With the same expert size, the relative computation cost of routing for different models is the same as shown in Appendix~\ref{sec:relative-cost}.


\begin{figure*}[t]
    \centering
    \includegraphics[width=\linewidth]{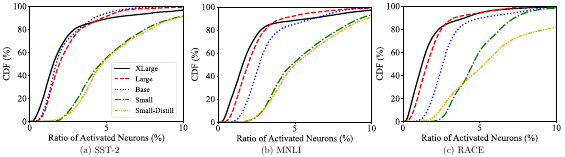}
    \caption{CDF of the ratio of activated neurons for each input with different models on three datasets.
    }
    \label{fig:scale-activation}
\end{figure*}

\begin{figure*}[t]
    \centering
    \includegraphics[width=\linewidth]{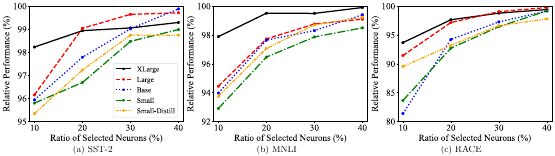}
    \caption{Relative performance of MoEfied models with different sizes on three datasets. Dynamically selecting $10\%$ to $20\%$ neurons can recover nearly $98\%$ original performance for large models such as T5-XLarge.}
    \label{fig:scale-performance}
\end{figure*}

\textbf{Expert Selection.} Besides Similarity Selection and MLP Selection, we also implement Random Selection, where we treat each expert as a collection of vectors with $d_{model}$ dimension and randomly select one of them as the expert representation. For Random Selection and Similarity Selection, the computation complexity for routing is O($kd_{model}$). For MLP Selection, we use a two-layer feed-forward network as the architecture. The input dimension is $d_{model}$, the intermediate dimension is $k$, and the output dimension is $k$. The non-linear activation function is $\tanh(\cdot)$. Its computation complexity is O($kd_{model} + k^2$). Compared to the computation complexity of FFNs of the original model, O($d_{model}\cdot d_{ff}$), the computation cost of routers is ignorable because $k$ is much smaller than $d_{ff}$. For example, $k$ is $128$ and $d_{ff}$ is $4096$ for T5-Large. 
For the training of our MLP routers, we adopt cross-entropy as the training objective and use the Adam optimizer with the learning rate of $10^{-2}$. The batch size is set to $512$ and the number of epochs is set to $10$. We sample nearly $500$ thousand input representations from the training data and split them into the training and development sets with the ratio of $9:1$. Note that we only use the activation information as supervision. The training time of each FFN is about several minutes on a single GPU.

\subsection{MoEfy ReLU-based Models}

In this subsection, we evaluate MoEfication on different T5 models. We consider two factors: the model size and whether the model is compressed. For the model size, we use five variants of T5~\cite{T5}, from T5-Small to T5-XLarge. For convenience, we directly use the scale names as the abbreviations. To investigate the influence of model compression, we compress T5-Large to T5-Small by classic knowledge distillation~\cite{KD}. 
Specifically, the teacher model is a fine-tuned T5-Large and the student model is a pre-trained T5-Small.
The distilled model is denoted by T5-Small-Distill. The expert construction and selection methods used here are Co-activation Graph Split and MLP Selection, which are proved to be the best combination in Section~\ref{sec:combinations}.

\begin{table}[]
    \centering
    \small
    \begin{tabular}{lrrr}
    \toprule
    Model         & SST-2 & MNLI & RACE \\
    \midrule
    Small         &    90.9   &   82.4   &   44.7   \\
    Small-Distill &   91.9    &   82.6   & 50.6     \\
    Base          &   94.0    &  86.4    &    71.7  \\
    Large         &   96.2    &    89.5  &  81.3    \\
    XLarge        &  96.9     &   90.5   &  85.6 \\
    \bottomrule  
    \end{tabular}
    \caption{Original Performance of different models on three downstream tasks. The model architecture is T5.}
    \label{tab:model-performance}
\end{table}

We report the performance of these models on three datasets, SST-2, MNLI, and RACE, in Table~\ref{tab:model-performance}. They are the representative datasets for single-sentence classification, sentence-pair classification, and reading compression, respectively. The original performance of PLMs grows as the model size grows, and knowledge distillation improves the performance of T5-small.

We first calculate the activation statistics of different models by inputting the training data of each dataset. The results are shown in Figure~\ref{fig:scale-activation}. 
From the figure, we have three observations. 
(1) The activations of these models are sparse. Different from the previous study on models trained with smaller datasets, where the activation ratios are range from $10\%$ to $50\%$~\cite{geva-etal-2021-transformer}\footnote{Since the activation ratios of a randomly-initialized model are around $50\%$, we guess these models do not make full use of their parameters.}, we find most inputs activate less than $10\%$ of the neurons.
(2) The activations of larger models are sparser than those of smaller models. For example, $80\%$ inputs only activate less than $3\%$ neurons in T5-XLarge while $40\%$ inputs activate more than $3\%$ neurons in T5-Small. 
(3) The sparsity is less related to distillation than the model size. The CDF curve of T5-Small-Distill is close to that of T5-Small.

Then, we compare the performance of MoEfied models with different sizes and ratios of selected neurons and report the results in Figure~\ref{fig:scale-performance}. To measure the performance of MoEfication, we calculate the relative performance of the MoEfied model to the original model. From the figure, we have four observations. (1) MoEfication works well with all models on all three datasets. MoEfied models use $10\%$ to $30\%$ of FFN parameters while maintaining over $95\%$ original performance. (2) The larger models can use fewer neurons to recover the original performance. For example, T5-XLarge achieves nearly $98\%$ relative performance on SST-2 and MNLI with $10\%$ neurons while T5-Small achieves the same results with $30\%$ to $40\%$ neurons. This result is consistent with the activation statistics, that is, larger models are sparser. We can expect that MoEfication can provide better efficiency with super large models. (3) Difficult tasks require models to select more experts to maintain the performance. From Table~\ref{tab:model-performance}, we can see that the accuracy of RACE is much lower than the other two tasks, and hence we think RACE is more difficult. Correspondingly, the relative performance with $10\%$ neurons on RACE is also lower than those on the other tasks. (4) MoEfication works similarly on T5-Small and T5-Small-Distill, which indicates that MoEfication can work with knowledge distillation for more efficient inference.

\begin{figure}[t]
    \includegraphics[width=\linewidth]{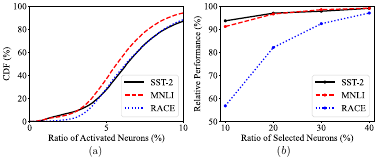}
    \caption{(a) CDF of the ratio of activated neurons in BERT-Large on SST-2, MNLI, and RACE. (b) Relative performance of MoEfied BERT-Large.}
    \label{fig:bert-large}
    \vspace{-1em}
\end{figure}

\subsection{MoEfy GeLU-based Models}

In addition to using ReLU as the activation function, many PLMs use GeLU~\cite{GeLU}, including BERT~\cite{BERT} and GPT~\cite{GPT-3}. In this subsection, we study whether BERT, which is the most representative GeLU-based model, can be MoEfied. Considering that GeLU gives negative inputs small activations instead of 0, we first transform a GeLU-based BERT into a ReLU-based BERT, and then MoEfy the ReLU-based model. Specifically, we initialize a ReLU-based BERT using the pre-trained parameters of a BERT-Large\footnote{\url{https://catalog.ngc.nvidia.com/orgs/nvidia/models/bert_pyt_ckpt_large_pretraining_amp_lamb}} and train the ReLU-based BERT on the pre-training corpus for the adaptation of the change of activation functions. In this work, we use the pre-training framework provided by NVIDIA\footnote{\url{https://github.com/NVIDIA/DeepLearningExamples}} and keep all hyper-parameters unchanged. Wikipedia and Bookcorpus are used as the pre-training corpus. In the experiments, after 400 optimization steps, the pre-training loss is close to that of the original model. Hence, the adaptation cost is much smaller than the pre-training cost (about 10000 steps). Meanwhile, the downstream performance of the ReLU-based model is comparable to the original model ($93.1$ v.s $93.5$ on SST-2 and $84.8$ v.s $85.2$ on MNLI). Based on this ReLU-based BERT-Large, we study the sparse activation phenomenon and the effect of MoEfication and report the results in Figure~\ref{fig:bert-large}.

From this figure, we have two observations: (1) The sparse activation phenomenon still exists in BERT. For example, more than 80\% of inputs activate less than 10\% of neurons. It reveals the generality of the sparse activation phenomenon in pre-trained Transformers. It will be an interesting direction to explain this phenomenon empirically or theoretically in the future. (2) MoEfication also archives good performance on BERT. For example, selecting $30\%$ to $40\%$ of neurons can recover $97\%$ performance. Since the activation of BERT is slightly denser than that of T5, it requires more neurons to recover most performance.

\begin{table}[t]
    \small
    \centering
    \setlength{\tabcolsep}{5pt}{
    \begin{tabular}{l|l|rrr}
    \toprule
       Construction    & Selection   & SST-2 & MNLI & RACE\\
    \midrule
    -                      & -           &    96.2   &   89.5     &    81.3 \\
    \midrule
    \multirow{4}{*}{Random}        & Groundtruth &   95.9    &    87.3    &   80.0     \\
    & Random      &    65.9   &   36.3     &  29.2      \\
    & Similarity       &    90.3   &   75.9     &  56.7        \\
    & MLP   &    \underline{94.1}   &   \underline{84.1}     &   \underline{75.0}       \\
    \midrule
    & Groundtruth &   95.5    &    88.8    &    80.9      \\
    Parameter & Random      &    70.6   &  36.4      &    41.8     \\
    Clustering & Similarity       &   86.7    &    66.3    &    63.6    \\
    & MLP   &   \underline{\textbf{95.9}}    &     \underline{\textbf{87.5}}   &    \underline{78.7}    \\
    \midrule
    & Groundtruth &   96.3    &    89.1    &       80.8   \\
    Co-Activation & Random      &   85.3    &     68.5   &   54.7     \\
    Graph & Similarity       &   92.2    &    81.4    &   71.0    \\
    & MLP   &   \underline{95.4}    &    \underline{\textbf{87.5}}    &     \underline{\textbf{79.0}}  \\
    \bottomrule
    \end{tabular}}
    \caption{Comparisons of different combinations of expert construction and selection methods using T5-Large. The first row is the original performance. The best results in each group are \underline{underlined} and the best results on each dataset are in \textbf{boldface}.
    }
    \label{tab:strategy}
\end{table}

\subsection{Comparisons of MoEfication Strategies}
\label{sec:combinations}

To find the most effective MoEfication strategy, we evaluate different combinations of expert construction and selection methods. We use T5-Large and also set the ratio of selected neurons to $20\%$. The results are shown in Table~\ref{tab:strategy}. From the table, we have two observations:

(1) For expert construction, Co-activation Graph Split is the best method according to the overall performance. Compared to the other two methods, Co-activation Graph Split directly uses the co-activation information to group the neurons activating simultaneously into the same expert.


(2) For expert selection, the performance of Groundtruth Selection is close to that of the original model, which indicates that $20\%$ parameters of FFNs are sufficient to achieve good performance on T5-Large. Meanwhile, MLP Selection is the best expert selection method and can work well with both Parameter Clustering Split and Co-activation Graph Split.

\section{Analysis}

In this section, we analyze the efficiency and routing patterns of MoEfied models.

\begin{table}[]
    \centering
    \begin{tabular}{lrrr}
    \toprule
    Ratio  & FLOPS & CPU  & GPU  \\
    \midrule
    50.0\%   & 1.50  & 1.43 & 1.15 \\
    25.0\%   & 2.00     & 1.98 & 1.20 \\
    12.5\% & 2.40   & 2.28 & 1.47 \\
    \bottomrule
    \end{tabular}
    \caption{Speedup of FLOPS, CPU and GPU with different ratios of selected neurons.}
    \label{tab:speedup-performance}
\end{table}

\subsection{Efficiency Improvement}

In this subsection, we show the efficiency improvement brought by MoEfication. We synthesize a batch of sequences with the input and output lengths of 64 and evaluate T5-Large on the data. To comprehensively show the efficiency improvement, we report the relative speedup based on FLOPS, CPU, and GPU in Table~\ref{tab:speedup-performance}. The FLOPS is estimated according to the statistics provided by~\citet{GPT-3}. The results of CPU and GPU are tested on an Intel Broadwell CPU and an NVIDIA Tesla V100 GPU, respectively.

From this table, we have three observations: (1) MoEfication can significantly reduce the total FLOPS, such as 2x speedup in the ratio of 25\%. Meanwhile, the speedup on CPU is close to that on FLOPS. Considering that CPU is widely used for model inference in real-world scenarios, MoEfication is practical for the acceleration of various NLP applications. 
(2) The smaller the ratio, the smaller the gain. For example, the gain of halving 25\% (to 12.5\%) is 1.2x while the gain of halving 50\% (to 25\%) is 1.3x. Although the FLOPS reduction of feed-forward networks is linear in the ratio, the cost of attention networks is unchanged and becomes the bottleneck. Hence, 20\% is a good ratio, which can have a significant speedup~(2x) and maintain most performance. 
(3) Since some of the operations of MoE cannot be easily paralleled, the speedup on GPU is smaller than that on GPU. Recently, some packages such as FastMoE~\cite{fastmoe} and Deepspeed-MoE~\cite{deepspeedmoe} are working on paralleling the inference of MoE models on distributed computing platforms and already have some promising results. We believe the bottleneck of parallel computing in MoE models will be well solved in the future.

\begin{figure}
    \centering
    \includegraphics[width=\linewidth]{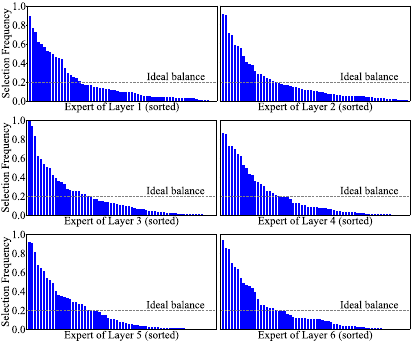}
    \caption{Selection Frequency of $64$ experts in each encoder layer of MoEfied T5-Small. The frequency of ideal balance selection is $0.2$ while the distribution is much unbalanced.}
    \label{fig:expert-distrib}
\end{figure}


\subsection{Routing Patterns}

In this subsection, we investigate the routing patterns of MoEfied models.
First, we count the selection frequency of each expert. Previous work introduces training objectives to ensure balance selection to make full use of model parameters~\cite{GShard,Switch-Transformer}. We report the results of the MoEfied T5-Small with $20\%$ experts on SST-2 in Figure~\ref{fig:expert-distrib}. From the figure, we observe that the frequency distribution of expert selection is much unbalanced. There are some commonly-used experts, whose frequencies are higher than $80\%$. Meanwhile, there are also some long-tail experts whose frequencies are lower than $10\%$.

\begin{figure}
    \centering
    \includegraphics[width=\linewidth]{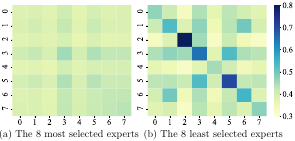}
    \caption{Input similarities between experts in the last encoder layer of MoEfied T5-Small. For the most selected experts, both the self-similarities and inter-similarities are low. For the least selected experts, the self-similarities are much higher than inter-similarities.}
    \label{fig:expert-sim}
\end{figure}

Then, we calculate the self-similarities and inter-similarities of inputs between experts by sampling $10,000$ inputs for each expert. 
We report the results of the last layer in Figure~\ref{fig:expert-sim}.
For the most selected experts, which are selected by most inputs, the self-similarities are close to the inter-similarities. For the least selected experts, the self-similarities are much higher than the inter-similarities, which suggests that the inputs of each expert have obvious cluster structure.

From these results, we can conclude the routing patterns of MoEfied models: there are some general experts, which can work for most inputs, and some input-specific experts, which are seldom used and may work in specific domains or tasks. This observation may inspire future work on training MoE models from scratch.



\section{Conclusion}


In this work, we verify that Transformer FFNs are naturally mixtures of experts and propose MoEfication, which utilizes the sparse activation phenomenon in FFNs to convert a normal model to its MoE version with the same parameters.
Experimental results show that MoEfied models can achieve comparable performance to the original models using only $10\%$ to $30\%$ of FFN parameters. Correspondingly, it significantly reduces the FLOPS of inference, e.g., 2x speedup with 20\% of FFN parameters.
Besides, by studying the routing patterns of MoEfied models, we find that there are general and input-specific experts, which may inspire future work on training MoE models.
We hope MoEfication can benefit real-world applications of PLMs with better efficiency and benefit the interpretation of the inner mechanism of FFNs.

\section*{Acknowledgement}

This work is supported by the National Key R\&D Program of China (No. 2020AAA0106502), Institute Guo Qiang at Tsinghua University, Beijing Academy of Artificial Intelligence (BAAI), and International Innovation Center of Tsinghua University, Shanghai, China. We thank Chenglei Si, Tianyu Gao and other members of THUNLP for their helpful discussion and feedback. Zhengyan Zhang conducted the experiments. Zhengyan Zhang, Yankai Lin, Zhiyuan Liu, and Peng Li wrote the paper. Maosong Sun and Jie Zhou provided valuable advices to the research.

\bibliography{anthology,custom}
\bibliographystyle{acl_natbib}

\newpage
\appendix



\section{MoEfication on Other Datasets}
\label{sec:other}

\begin{table*}[t]
        \small
        \centering
        \setlength{\tabcolsep}{5pt}{
        \begin{tabular}{l|cccccccccc|l}
        \toprule
        & MNLI & QNLI & QQP & RTE & SST-2 & MRPC & CoLA & STS-B & RACE & SQuAD 1.1 & \multicolumn{1}{c}{Avg.} \\
        \midrule
        Original    &   89.5   &  94.4    &  91.7   &  87.1   &   96.2  &   88.0   &   59.4   &   91.2/90.9  &   81.3   &    93.2 & 87.2  \\
        \midrule
        MoEfied &  87.5    &    93.2  &   90.2  &  86.4   &  95.4   &  87.5    &   55.5   &  90.6/90.3   &   79.0   &   92.2 & 85.7 (-1.5) \\
        \quad+GT &   89.1   &   94.1   & 91.4    &  86.4   &   96.3  &  88.3    &   58.8   &  90.9/90.8   &   80.8   &    93.2 & 86.9 (-0.3) \\
        \quad+Calib  &  88.7    &    93.6  &  91.3   &  87.5  &  96.2   &  89.3    &   59.4   &  91.0/90.6   &   79.9   &  92.3 & 86.9 (-0.3) \\
        \bottomrule  
        \end{tabular}}
        \caption{Results of T5-Large on GLUE benchmark and two QA datasets. The last row reports the differences between the original model and MoE+Calib. MoEfied models with parameter calibration achieve comparable performance to original models.}
        \label{tab:res}
        \vspace{-0.5em}
    \end{table*}

For text classification, we use GLUE benchmark~\cite{GLUE}, including MNLI-matched~\cite{williams2018broad}, QNLI~\cite{SQuAD}, QQP\footnote{\url{https://data.quora.com}}, RTE~\cite{dagan2006pascal}, SST-2~\cite{socher2013recursive}, MRPC~\cite{dolan2005automatically}, CoLA~\cite{warstadt2018neural}, and STS-B~\cite{giampiccolo2007third}. For reading comprehension, we use SQuAD~\cite{SQuAD} and RACE~\cite{RACE}, which are the representative datasets for span extraction and multi-choice QA, respectively. We report the results on their development sets. For MNLI, QNLI, QQP, RTE, SST-2, MRPC, RACE, we use accuracy as the metric. For CoLA, we use matthews correlation coefficient as the metric. For STS-B, we use pearson and spearman correlation as the metrics. For SQuAD, we use F1 score as the metric.

We evaluate MoEfication on several downstream natural language understanding tasks with T5-Large. 
The ratio of selected neurons is set to $20\%$, which is sufficient for T5-Large as show in Figure~\ref{fig:scale-activation}. In practice, there is still a gap between the performance of MoEfied models and that of original models because selected experts cannot cover all positive neurons with a limited computation budget. Hence, the outputs of MoEfied models will be slightly different from those of original models. To calibrate MoEfied models, we further fine-tune the models on the training set, namely parameter calibration. Considering that current routers are based on the first layers of FFNs ($\mW_1$ and $\vb_1$), we only optimize the second layers of FFNs ($\mW_2$ and $\vb_2$) to ensure routers can also work well after fine-tuning. We use a small learning rate of $10^{-7}$ for calibration. The other hyper-parameters remain the same as fine-tuning. The results are shown in Table~\ref{tab:res}. MoEfied refers to the combination of Co-activation Graph Split and MLP Selection. MoEfied+GT refers to the combination of Co-activation Graph Split and Groundtruth Selection. MoEfied+Calib is the calibrated version of MoEfied. To calculate the average performance, we also include SST-2, MNLI, and RACE.

We observe that MoEfication introduces small performance loss (about 1.5\% on average) with an 80\% reduction of the computation cost in FFNs. Meanwhile, calibration can effectively deal with the issue of the precision errors brought by MoEfication. For example, MoEfied+Calib improves MoEfied by nearly 4\% on CoLA and achieves the same average performance as MoEfied+GT.

\section{Activation Statistics before Fine-tuning}

\begin{figure}[h]
\includegraphics[width=\linewidth]{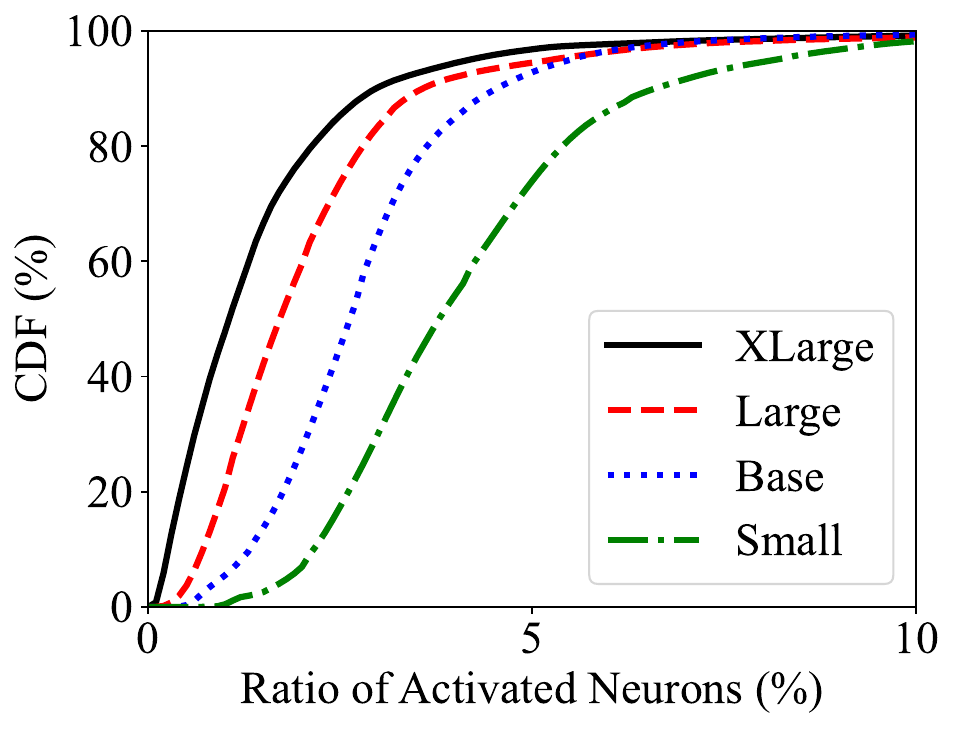}
\caption{CDF of the ratios of activated neurons for each input with different models before fine-tuning.}
\label{fig:pretrain-activation}
\end{figure}

We count the activation statistics of PLMs before fine-tuning on the pre-training data containing about $50,000$ input tokens. The results are shown in Figure~\ref{fig:pretrain-activation}. We observe that PLMs before fine-tuning also have the sparse activation phenomenon and fine-tuning brings little change.

Then, we compare the activations of pre-trained models and those of fine-tuned models. We use the average ratio of activated neurons as the index. The results are shown in Table~\ref{fig:average-ratio}. We observe that fine-tuning increases the average activation ratio for most models. The reason may be that different neurons start to learn the same task-specific patterns during fine-tuning. Interestingly, the increase on RACE is smaller than that on the other datasets. Since RACE is more difficult than the other datasets, there should be more task-specific patterns in RACE and less neurons learn the same patterns. Moreover, the pre-training task MLM requires more patterns than RACE so the ratios of MLM are lowest.

\begin{table}[h]
\centering
\begin{tabular}{lrrrr}
\toprule
      & Small & Base & Large & XLarge \\
\midrule
MLM   & 4.18  & 2.85 & 2.17  & 1.52   \\
\midrule
SST-2 & 5.53  & 2.24 & 2.50  & 2.46   \\
MNLI  & 5.59 & 3.25 & 2.44  & 2.45   \\
RACE  & 4.94 & 3.08 & 1.98  & 1.79  \\
\bottomrule
\end{tabular}
\caption{Average ratio of activated neurons for each input. MLM represents the pre-trained models with masked language modeling. SST-2, MNLI, RACE represent the fine-tuned models on each dataset.}
\label{fig:average-ratio}
\end{table}

\section{Results of Graph Partition}

\begin{figure}[h]
\centering
        \includegraphics[width=0.8\linewidth]{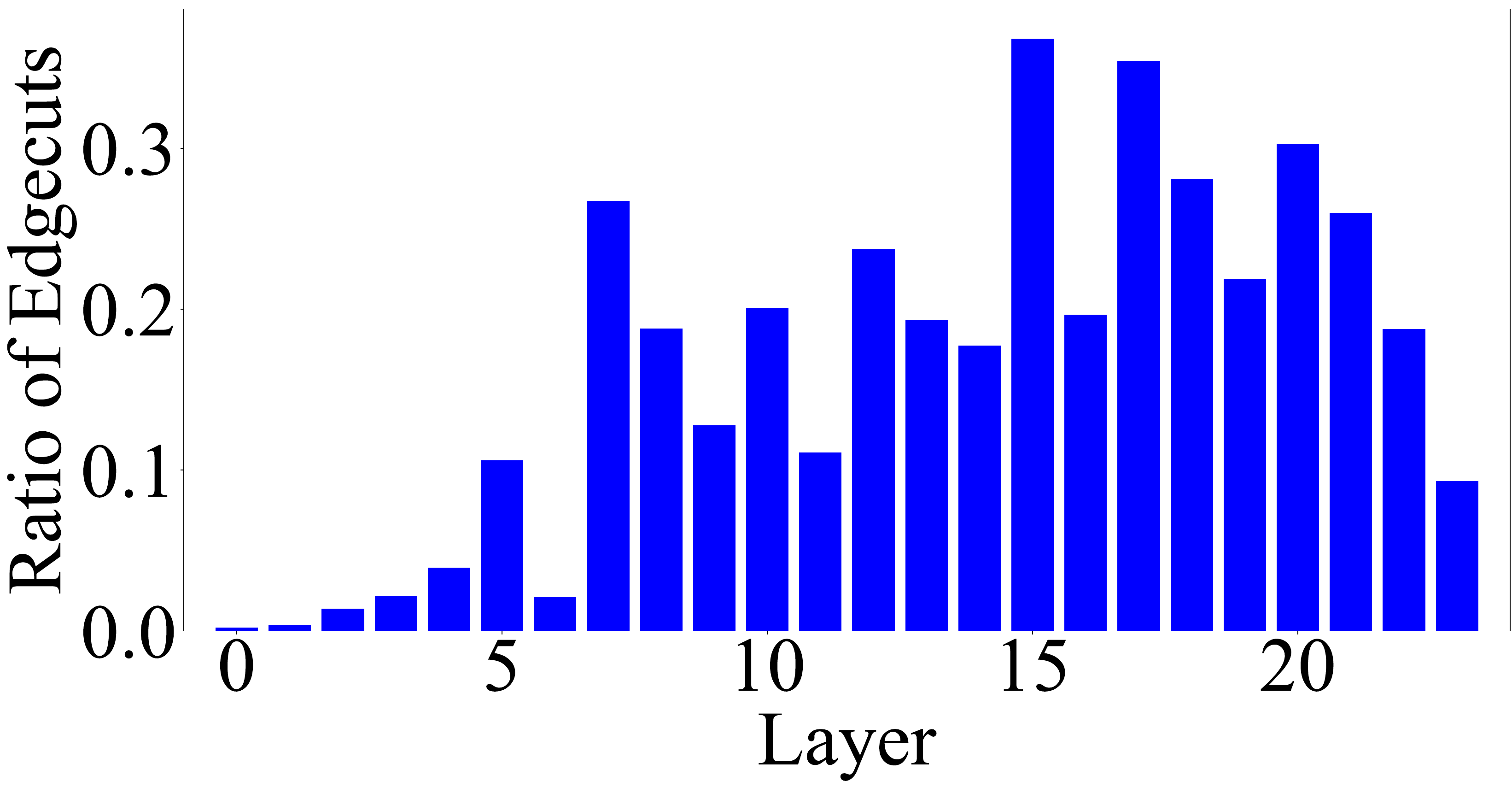}
        \caption{Ratio of edgecuts in different layers.}
        \label{fig:partition}
\end{figure}

Co-activation Graph Split achieves good performance in expert construction. Here, we study whether the co-activation graph is suitable for partitioning. We report the results of graph partition of T5-Large on SST-2 in Figure~\ref{fig:partition}. Smaller ratios of edgecuts, which straddle partitions, mean that more co-activation pairs are included in experts. We only report the results of encoder layers because all ratios of decoder layers are smaller than $0.001$. From this figure, we can see that the overall ratio is small and these graphs are suitable for partitioning.

\section{Accuracy of MLP Selection}

\begin{figure}[h]
\centering
        \includegraphics[width=0.8\linewidth]{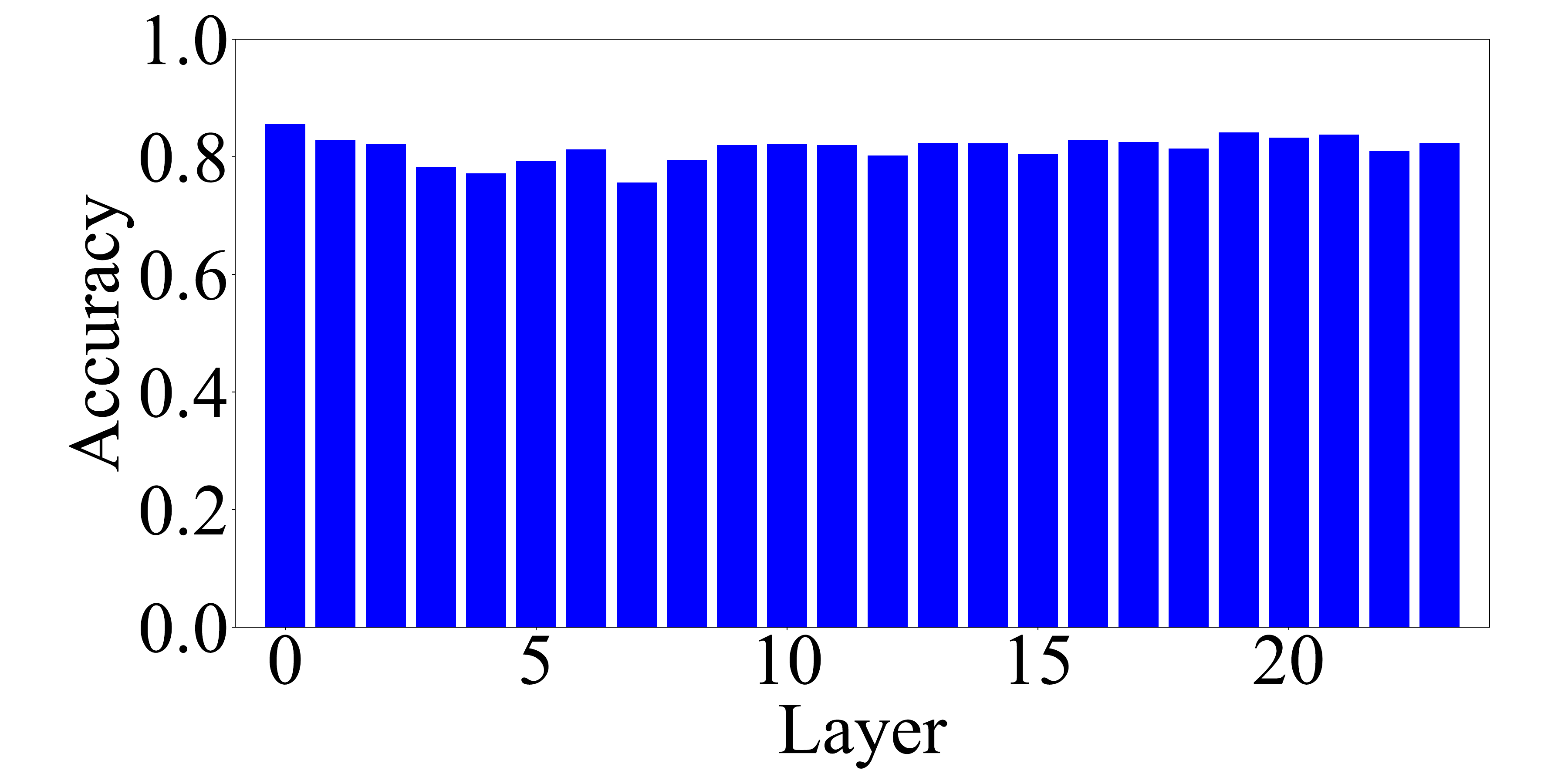}
        \caption{Accuracy of MLPs of encoder layers.}
        \label{fig:mlp_acc_enc}
\end{figure}

\begin{figure}[h]
        \centering
        \includegraphics[width=0.8\linewidth]{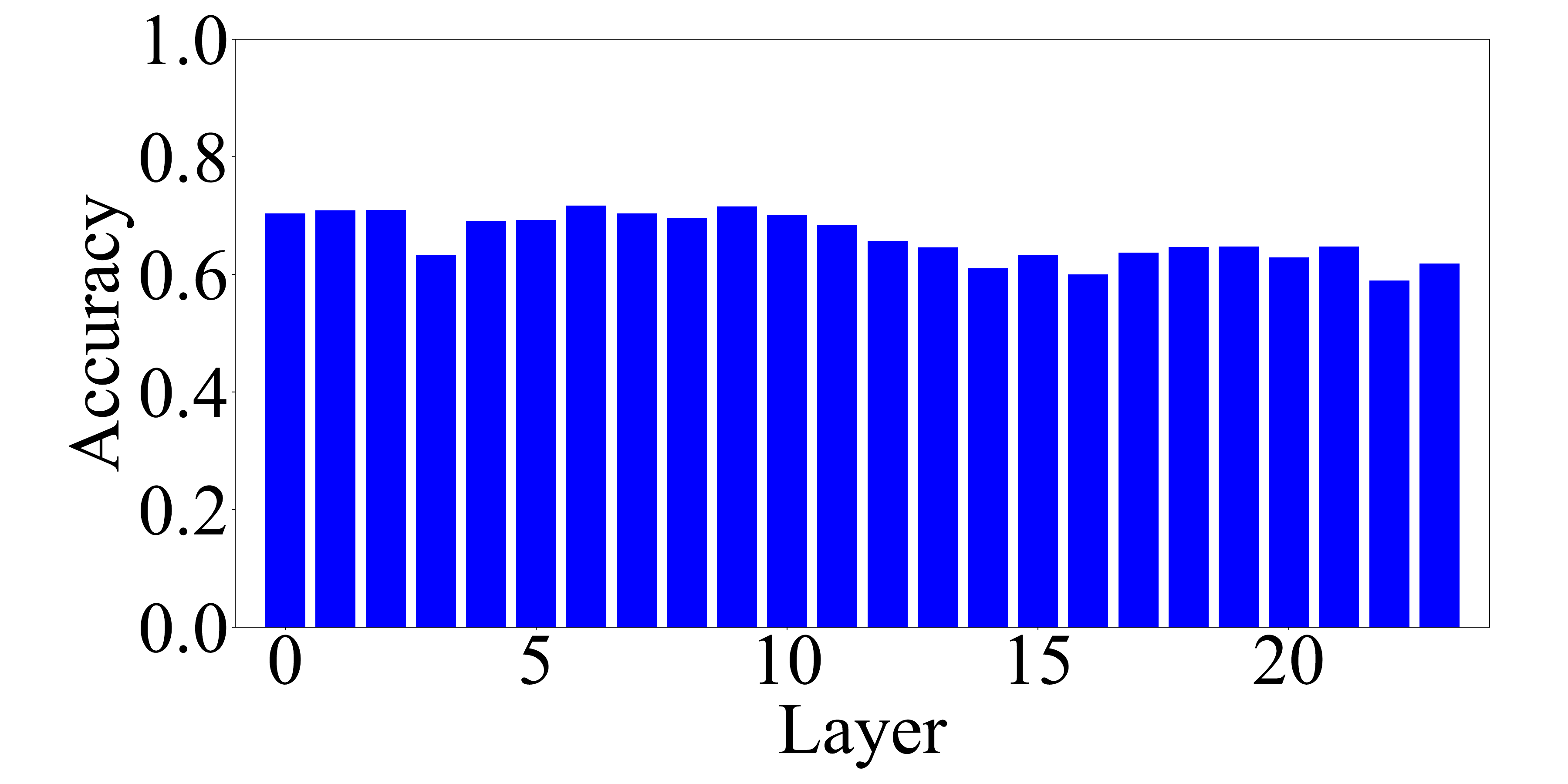}
        \caption{Accuracy of MLPs of decoder layers.}
        \label{fig:mlp_acc_dec}
\end{figure}

MLP selection trains MLPs to fit the groundtruth selection. In this part, we report the accuracy of MLPs in T5-Large fine-tuned on SST-2. The results are shown in Figure~\ref{fig:mlp_acc_enc} and~\ref{fig:mlp_acc_dec}. The overall accuracy of the encoder is about $0.8$ and the overall accuracy of the decoder is about $0.7$.

\section{Relative Cost of Routing}
\label{sec:relative-cost}

In this work, we set the number of neurons in each expert to $32$. Then, the number of experts in each layer $k$ is $\frac{d_{ff}}{32}$. In most Transformer models, $d_{ff} = 4d_{model}$. The computation complexity of Similarity Selection for each input is
\begin{equation}
    O(kd_{model})= O(\frac{d_{model}^2}{8}).
\end{equation}
The computation complexity of FFNs for each input is
\begin{equation}
    O(d_{model}\cdot d_{ff}) = O(4d_{model}^2).
\end{equation}
Then, the relative cost of routing to that of FFNs is constant for different models. It is also similar to MLP Selection.

\section{Graph Partitioning Algorithm}
\label{sec:partition}

The goal of graph partitioning is to divide a graph into several sub-graphs where the number of edges crossing sub-graphs is minimized. 
In this work, we use the graph partitioning algorithm proposed by~\citet{metis}.
The graph partitioning algorithm consists of three phases: coarsening phase, partitioning phase, and refinement phase.
(1) In the coarsening phase, we create new super nodes by grouping nodes that are highly connected together. For example, if the weight of the edge between two nodes is large, these two nodes will be grouped together. In the setting of coarsening co-activation graphs studied in this work, two neurons that often activate simultaneously will be treated as a new super neuron. (2) In the partitioning phase, we start with an initial bipartition of the super node graph and then iteratively search for super nodes from each part of the graph, such that swapping them leads to a partition with a smaller number of crossing edges. To divide a graph into $k$ parts, we need $\log k$ rounds of bipartition. (3) In the refinement phase, we project super nodes to the original nodes and then continue to iteratively swap nodes to reduce the number of crossing edges.

\section{Comparisons with Model Pruning}
\label{sec:model-prune}

\begin{figure}
  \centering
  \includegraphics[width=0.6\linewidth]{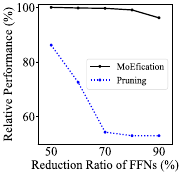}
  \caption{Comparison between MoEfication and  model pruning.}
  \label{fig:pruning}
\end{figure}

Based on the fine-tuned T5-Large on SST-2, we compare MoEfication with model pruning, which omits the weight having small values. The results are shown in Figure~\ref{fig:pruning}. We observe that model pruning significantly degrades the performance.
However, MoEfication achieves good performance by selectively activating parts of the network according to input.

\begin{table}[t]
    \small
    \centering
    \begin{tabular}{ll}
    \toprule
    Model & MLM Loss \\
    \midrule
    MoE Pre-training     &       3.09 \\
    \midrule
    Standard Pre-training   &   2.88 (-0.21)       \\
    \quad +MoEfication      &    3.02 (-0.07)     \\
    \quad +GT      &   2.95 (-0.14)      \\
    \bottomrule
    \end{tabular}
    \caption{Comparisons of MoE models pre-trained from scratch and modified by MoEfication. We report the MLM loss on the validation set. Standard pre-training with MoEfication is better than pre-training a MoE model from scratch.}
    \label{tab:moe-scratch}
    \vspace{-0.5em}
\end{table}

\section{MoEfication vs. MoE pre-training}

In this subsection, we compare the performance of two kinds of MoE models. The first one is pre-trained from scratch. The second one is transformed from a standard model by MoEfication. For fair comparisons, we pre-train one MoE model and one standard model with the same model size from scratch using WikiText-103~\cite{wikitext-103}. The pre-training objective is masked language modeling (MLM). The model architecture is the same as T5-Small. For pre-training, we use the batch size of $4096$, the learning rate of $0.01$, the maximum sequence length of $512$, and the Adam optimizer. The number of experts is set to $64$ and the router will select $32$ of them for a single input.

We report the MLM loss on the validation set in Table~\ref{tab:moe-scratch}. From the table, we have two observations. (1) The loss of the standard pre-trained model is lower than that of the pre-trained MoE model. We guess that the optimization of MoE models is difficult than that of the standard models because of the restricted selection of MoE models. (2) MoEfied models achieve better performance than the pre-trained MoE model. It indicates that pre-training a standard model then conducting MoEfication can be a better option than pre-training an MoE model from scratch. 

\end{document}